\documentclass{article}

\usepackage{rldm}
\usepackage{natbib}
\usepackage[utf8]{inputenc} 
\usepackage[T1]{fontenc}    
\usepackage{hyperref}       
\usepackage{url}            
\usepackage{booktabs}       
\usepackage{amsfonts}       
\usepackage{nicefrac}       
\usepackage{microtype}      
\usepackage{graphicx}
\usepackage{amsmath}
\usepackage[ruled]{algorithm2e}
\usepackage{subcaption}
\usepackage{float}

\setcitestyle{authoryear,open={(},close={)}}

\SetKwComment{Comment}{/* }{ */}

\title{Learning to Query Internet Text for \\Informing Reinforcement Learning Agents}

\author{
  Kolby Nottingham \\
  UC Irvine \\
  Irvine, CA 92697 \\
  \texttt{knotting@uci.edu} \And
  Alekhya Pyla \\
  UC Irvine \\
  Irvine, CA 92697 \\
  \texttt{apyla@uci.edu} \And
  Sameer Singh \\
  UC Irvine \\
  Irvine, CA 92697 \\
  \texttt{sameer@uci.edu}  \And
  Roy Fox \\
  UC Irvine \\
  Irvine, CA 92697 \\
  \texttt{royf@uci.edu}
}

\begin{document}

\maketitle

\begin{abstract}
    Generalization to out of distribution tasks in reinforcement learning is a challenging problem. One successful approach improves generalization by conditioning policies on task or environment descriptions that provide information about the current transition or reward functions. Previously, these descriptions were often expressed as generated or crowd sourced text. In this work, we begin to tackle the problem of extracting useful information from natural language found in the wild (e.g. internet forums, documentation, and wikis). These natural, pre-existing sources are especially challenging, noisy, and large and present novel challenges compared to previous approaches. We propose to address these challenges by training reinforcement learning agents to learn to query these sources as a human would, and we experiment with how and when an agent should query. To address the \textit{how}, we demonstrate that pretrained QA models perform well at executing zero-shot queries in our target domain. Using information retrieved by a QA model, we train an agent to learn \textit{when} it should execute queries. We show that our method correctly learns to execute queries to maximize reward in a reinforcement learning setting.
\end{abstract}


\section{Introduction}

Reinforcement learning agents have recently begun to be deployed in real world applications, but these systems still tend to be repetitive tasks with well defined state spaces. Better generalization is needed for applications that interact with the more dynamics parts of the real world. Towards this end, researchers have been attempting to assist reinforcement learning agents using natural language, an abundant type of data that is easy for humans to provide and understand. In this work we explore using domain data found in the wild to help agents learn better in given domain.

Using text data found in the wild is important for scaling reinforcement learning methods to real world applications \citep{luketina2019survey}. Some of the most abundant sources of informative language in the wild are data specific to a domain but independent of any particular task in that domain (e.g. internet forums, wikis, and other documentation). This type of data tends to be noisy, large, and filled with information irrelevant for the current task. While most previous work informing reinforcement learning agents with natural language uses text that is generated from rules and templates \citep{chevalier2018babyai,zhong2019rtfm} or crowdsourced for a specific task \citep{Chen_2019_CVPR,shridhar2020alfred,hanjie2021grounding}, we use challenging task-independent domain text found in the wild.

We propose teaching reinforcement learning agents to actively query natural language sources for information to help in its current task. Humans often query the internet to help them generalize to new tasks. We identify three major components to teaching artificial agents to do the same. Those are \textit{how}, \textit{when}, and \textit{what} to query. We address the first two in this work and leave the latter for future work. 

To address the \textit{how}, we use a pretrained QA model to extract structured representations from text in a zero-shot setting. Next, we teach an agent \textit{when} to query by adding a query action in the agent's action space. Our agent successfully retrieves relevant information from text and only does so when required for the current task. We use the nethack learning environment for our experiments along with the game's online wiki pages \citep{kuttler2020nethack}.


\section{Related Work}

\subsection{Task \& Domain Independent Language}

Language is becoming a popular method for informing reinforcement learning agents. This is partly motivated by recent advancements in powerful pretrained language models. Language models have been used in reinforcement learning applications to process text data such as natural language instructions \citep{hill2020human} or observations grounded in text \citep{ammanabrolu2020avoid,yin2020learning,li2022pre}. Language models are generally found to increase generalization capability.

Other work, more closely related to ours, has attempted to impart common sense to reinforcement learning agents from the language model itself \citep{dambekodi2020playing, ahn2022can}. Because lanuage models are trained on vasts amount of data that are independent of both an agent's target domain and task, this approach limits knowledge to information that the language model was repeatedly exposed to during training. This is usually referred to as common sense knowledge and doesn't provide the same type of knowledge found in domain dependent language.

\subsection{Task \& Domain Dependent Language}

Language that is both task and domain dependent is specific to the current task and can provide information about the agent's current reward and transition functions. For example, \cite{zhong2019rtfm} and \cite{hanjie2021grounding} train agents conditioned on descriptions on the current environment and goal. These descriptions are gathered using generation templates and crowdsourcing respectively.

In an effort to work towards methods that scale, we choose to use language that is found in the wild. We found that language found in the wild is often domain dependent but usually task-independent. We believe that focusing on this type of language data will lead to valuable research as it is more applicable than domain independent data but more available in the wild than task dependent data. 

\section{Learning How to Query}
\label{sec:how}

Text in the wild often consists mostly of information that is irrelevant to our current task. Also, there are typically too few instances found in the data to train models from scratch. Thus, our goal is to suggest a method for extracting specific information from language with little to no training data. To this end we propose using pretrained QA models prompted with task specific questions.

\subsection{Zero-shot Performance}

\begin{table}[t]
\small
\setlength{\tabcolsep}{2.75pt}
\center
\begin{tabular}{lcccccccc}
\toprule
& \multicolumn{4}{c}{\bf Resistance}  & \multicolumn{4}{c}{\bf Attack} \\
\cmidrule(lr){2-5}
\cmidrule(lr){6-9}
    & recall & precision & F1 & IOU & recall & precision & F1 & IOU \\
\midrule
    Keyword & 0.62 & 0.64 & 0.63 & 0.46 & \textbf{0.98} & 0.15 & 0.26 & 0.15 \\
    UnifiedQA & \textbf{0.81} & \textbf{0.72} & \textbf{0.76} & \textbf{0.61} & 0.74 & \textbf{0.53} & \textbf{0.62} & \textbf{0.45} \\
\bottomrule
\end{tabular}
\caption{Metrics for predicting monster resistance and damage types from nethack wiki pages. The keyword baseline searches the page for the resistance and attacks type directly. The UnifiedQA method searches the QA model output for the resistance and damage keywords.}
\label{table:qa}
\end{table}

We analyze the ability of the QA Model to perform in a zero-shot setting by testing its ability to recover monster resistance and damage types from the nethack wiki (\url{https://nethackwiki.com/}). We label a small evaluation set of monster data for 98 pages. Monsters are labeled as having a set of resistance and attack types. There are eight potential resistances and 17 potential attack types. Table \ref{table:qa} shows the performance of a three billion parameter pretrained UnifiedQA model \citep{khashabi2020unifiedqa} compared with a simple keyword search baseline. 

For the resistance task we prompted UnifiedQA with the contexts ``What is it resistant to?'' and ``What is it's resistance?''. For the attack task we prompt the model with ``What attack does it do?'' and ``What type of damage does it do?''. We check the joint generated output for the questions for the set of resistance and attack types. Our reported metrics compare this predicted set with the labeled set.

The attack type task was the more of the two because of the greater number of attribute classes and because attack types of other monsters were commonly mentioned on a monster's wiki page. This made precision on this task suffer, but, for the most part, the UnifiedQA model was able to filter these irrelevant mentions of attack types.

\subsection{Representation Generalization}

\begin{figure}
     \centering
     \begin{subfigure}{0.4\textwidth}
         \centering
         Training Task
         \includegraphics[width=\textwidth]{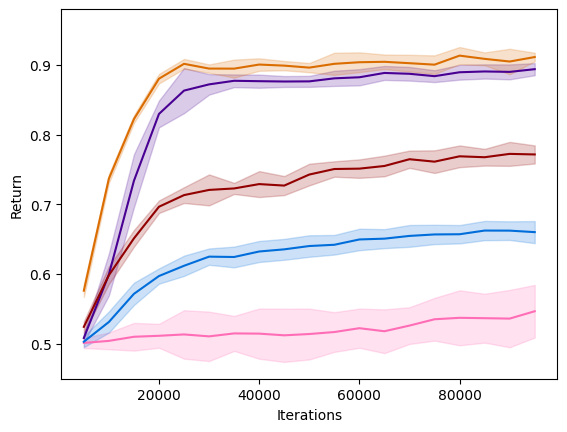}
     \end{subfigure}
     \begin{subfigure}{0.38\textwidth}
         \centering
         Evaluation Task
         \includegraphics[width=\textwidth]{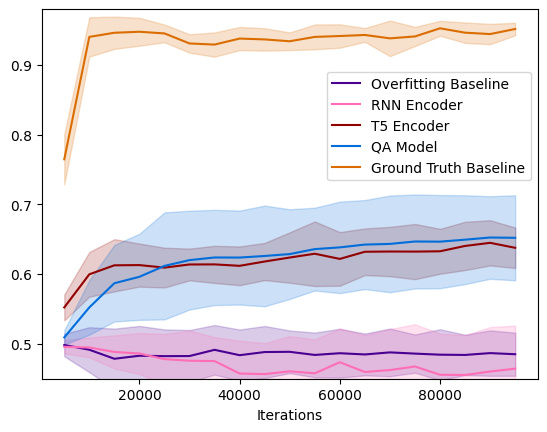}
     \end{subfigure}
    \caption{Returns for each method on the training and evaluation tasks. We perform an 80:20 training evaluation split over the nethack monsters. This way, the evaluation task only sees novel monsters. Error bars represent standard deviation over 10 random seeds.}
    \label{fig:res}
\end{figure}

To compare methods for extracting information from natural language data, we use a contextual bandit problem based on the nethack learning environment \citep{kuttler2020nethack} to compare the below text representation methods. 

\textbf{No External Source Baseline:} This method trains a value function on a one-hot encoding of the state with no additional knowledge. It is expected to overfit.

\textbf{RNN Encoding:} We train a single layer LSTM to encode the provided language data. The final hidden layer of the LSTM is passed to the value function and both are trained end to end.

\textbf{LM Encoding:} This approach uses a pretrained LM with frozen weights to encode language data. The extracted text features is passed to the value function as a vector encoding.

\textbf{QA Model:} The QA approach defines a fixed set of questions to ask using the language data as context. The language data is then represented as a binary vector encoding the responses to each question. We write a set of questions manually to prompt the model.

\textbf{Ground Truth QA Baseline:} The ground truth baseline is identical to the QA model but with ground truth query responses rather than responses returned by the model. This method represents ideal generalization with the caveat that it may provide information that doesn't exist in the original external language source.

In our setup, the agent is given the goal of receiving a target resistance (fire, shock, sleep, poison, or cold) by consuming a monster corpse. It is then given a choice between two monsters (out of 388) along with the wiki pages each monster. One of the monsters is guaranteed to confer the target resistance. The agent receives a reward of one if it chooses the monster that will confer the target resistance and a zero otherwise.

Figure \ref{fig:res} shows the results of our contextual bandit over 100,000 iterations. As expected the baselines both perform well on the training task while only the ground truth baseline generalizes to the evaluation task. Due to the extra monster identifying information contained in the LM encoded representation, it outperforms the QA model on the training task, but it does not generalize any better than the QA model on the evaluation task. The RNN struggles to encode the needed information from the wiki pages on both tasks. The only methods that do not overfit to the training data are the QA model and the ground truth baseline.

\section{Learning When to Query}
\label{sec:when}

\begin{table}[t]
\small
\setlength{\tabcolsep}{2.75pt}
\center
\begin{tabular}{lccc}
\toprule
    & Reward & Weapon Choice & Query Relevance\\
\midrule
    Baseline Agent & 2.16 & 0.55 &  \\
    Query Agent & 2.25 & 0.89 & 0.64\\
    Query+Explore Agent & \textbf{2.28} & \textbf{0.95} & \textbf{0.67} \\
\hline
    Oracle Agent & 2.34 & 0.98 & \\
\bottomrule
\end{tabular}
\caption{Performance of reinforcement learning agents in nethack. Reward is the number of monsters killed in an episode. Weapon choice is the percent of time that the agent chose to attack with the correct weapon for a corresponding monster. Query relevance is the percent of queries the agent makes that were significant in determining the correct weapon choice. The querying agent is compared to agents that have all important information without needing to query (Oracle) and an agent that is deprived of that information and cannot query (Baseline). Finally, we experiment with adding a special exploration policy specifically for queries (+Explore).}
\label{table:query}
\end{table}


In partially observable environments, a reinforcement learning agent is given an imperfect observation of the current environment state. Additional natural language data can be viewed as providing additional observations to an agent to help disambiguate between states. When actively querying language data, an agent should be able to learn to query only in states where the additional information will increase its expected return. 

We formulate querying as part of the reinforcement learning problem by adding a query action to the agent's action space. In our experiments, executing a query receives the same timestep penalty that all actions receive. This way the agent learns to only query when it is optimal to do so.

We test this method by designing custom nethack levels using the minihack library \citep{samvelyan2021minihack}. We spawn an agent with two choices of weapon in its inventory in a level full of four types of monsters. For two of these monsters, the choice of weapon the agent attacks with is significant because the monsters have resistances to one item or the other. For the other two, both weapons work equally well. We give the agent the ability to query the resistances of nearby monsters using the monster wiki pages and the QA method described in section \ref{sec:how}. The agent is rewarded each time it successfully slays a monster until it dies or the episode horizon is reached.

Table \ref{table:query} shows the results of our querying agent compared with baselines that either always or never have the monster resistance information. We report the reward along with how often the agent chooses to attack with the correct weapon and how often the agent's queries are for the monsters with different resistances. The agent should be able to learn to manage its inventory to always attack with the optimal weapon. It should also learn to only query the wiki when necessary, avoiding unnecessary queries about the monsters with identical resistances.

We found that the base querying agent took nearly eight times longer to learn to use resistance information in this task. To speed up this process we implement an exploration policy for taking the query action. At the start of training, the agent is forced to take the query action with 25\% probability. This probability decreases to zero over the first five million steps of training. Doing this helps the agent learn to utilize queries sooner. As shows in table \ref{table:query}, the query+explore agent converged in the same amount of time as the baseline and oracle agents and learned a better policy.

\section{Conclusion}

In this work, we teach a reinforcement learning agent how and when to query for additional information about its environment. We use natural language data found in the wild for the agent to query and discuss how to overcome the challenges that come from this type of data. We find that pretrained QA models are good at retrieving structured zero-shot representations of text. Additionally, reinforcement learning agents can learn when to query this information and when not to.

In future work, we're excited to see reinforcement learning agents with more advanced methods for querying. In addition, further work should explore what to query in the current context of the environment in addition to how and when.

\bibliography{queries}

\end{document}